\pdfoutput=1

\documentclass[11pt]{article}
\usepackage[]{acl}
\usepackage{times}
\usepackage{latexsym}
\usepackage[T1]{fontenc}
\usepackage[utf8]{inputenc}
\usepackage{microtype}
\usepackage{inconsolata}
\usepackage{graphicx}
\usepackage{xspace}

\usepackage{inconsolata}
\usepackage{kotex}
\usepackage[flushleft]{threeparttable} 
\usepackage{makecell}
\usepackage{graphicx}
\usepackage{booktabs}
\usepackage{multicol}
\usepackage{makecell}
\usepackage{amsmath}
\usepackage{geometry}
\usepackage{array}
\usepackage{tabularx}
\usepackage{amsmath}
\usepackage{longtable}
\usepackage{multirow}
\usepackage{footmisc}
\usepackage{placeins}
\usepackage{adjustbox}

\newcommand{\ours}{\textsc{LRAGE}}
\newcommand{\Ours}{\textsc{Legal Retrieval Augmented Generation Evaluation tool}}

\title{
    \adjustbox{valign=m}{
        \includegraphics[width=0.04\textwidth]{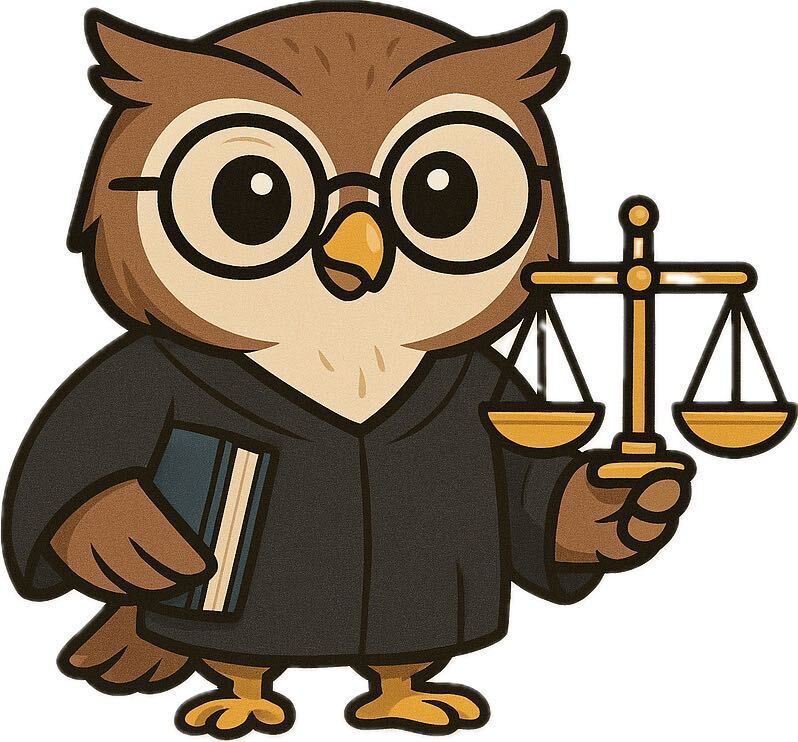}
    } LRAGE: Legal Retrieval Augmented Generation \\ Evaluation Tool
}

\author{
    Minhu Park$^{1,}$\thanks{Equal contribution.}
    \quad Hongseok Oh$^{1,}$\footnotemark[1]
    \quad Eunkyung Choi$^{1}$
    \quad Wonseok Hwang$^{1,2,}$\thanks{Corresponding author.} \\
    $^1$ University of Seoul \quad $^2$ LBOX
    \\ \texttt{\{alsgn2003, cxv0519, rmarud202, wonseok.hwang\}@uos.ac.kr}
}

\begin{document}
\maketitle

\begin{abstract}

Recently, building retrieval-augmented generation (RAG) systems to enhance the capability of large language models (LLMs) has become a common practice. Especially in the legal domain,  previous judicial decisions play a significant role under the doctrine of stare decisis which emphasizes the importance of making decisions based on (retrieved) prior documents. However, the overall performance of RAG system depends on many components: (1) retrieval corpora, (2) retrieval algorithms, (3) rerankers, (4) LLM backbones, and (5) evaluation metrics. Here we propose \ours, an open-source tool for holistic evaluation of RAG systems focusing on the legal domain. \ours\ provides GUI and CLI interfaces to facilitate seamless experiments and investigate how changes in the aforementioned five components affect the overall accuracy.
We validated \ours\ using multilingual legal benches including Korean (KBL), English (LegalBench), and Chinese (LawBench) by demonstrating how the overall accuracy changes when varying the five components mentioned above. The source code is available at \url{https://github.com/hoorangyee/LRAGE}.

\end{abstract}

\section{Introduction}

Recently large language models (LLMs) have demonstrated remarkable performance across a wide range of tasks. However, in expert domains--where average users struggle to assess the accuracy of an LLMs' responses--their performance remains limited due to a tendency to hallucinate \citep{dahl2024largelegalfiction,magesh2024lhallucinationfree}.

\begin{figure}[t!]
    \centering
    \includegraphics[width=1.0\columnwidth]{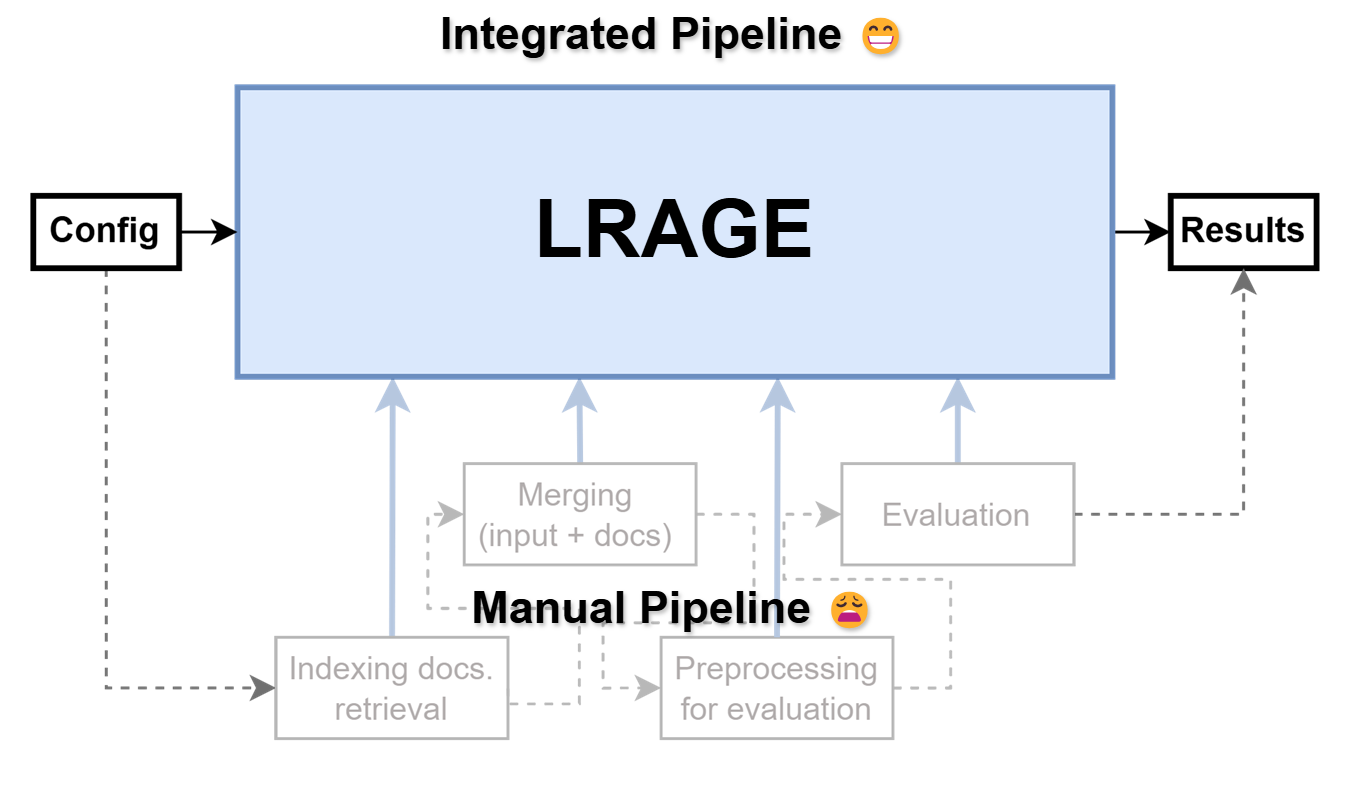}
    \caption{Comparison of conventional RAG evaluation pipeline (bottom) and the LRAGE framework (up) where each process is seamlessly integrated.}
    \label{fig: system}
\end{figure}

\begin{figure*}[htbp!]
    \centering
    \includegraphics[width=0.68\textwidth]{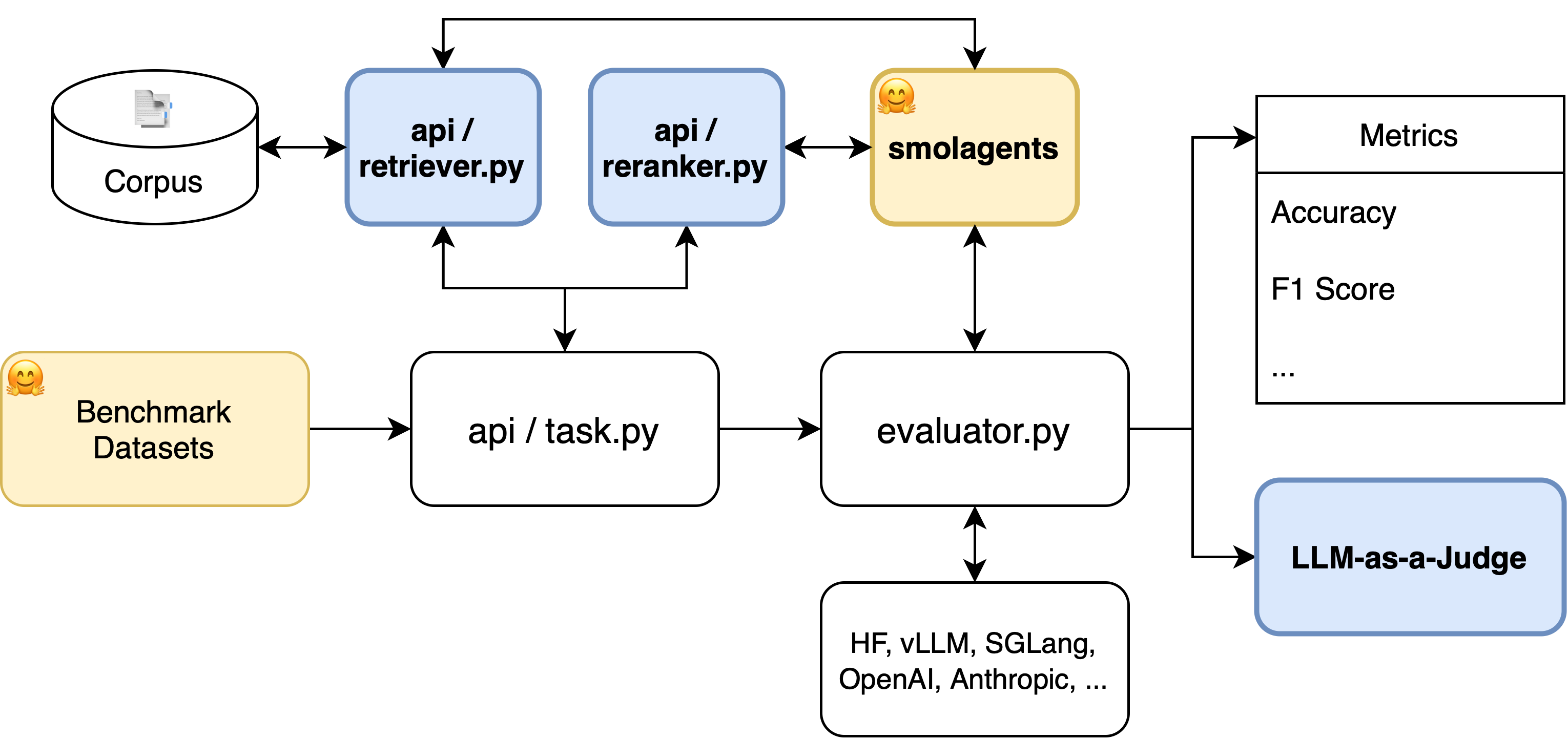}
    \caption{System diagram}
    \label{fig: system}
\end{figure*}
To address this limitation, it has become standard practice to employ Retrieval-Augmented Generation (RAG), which integrates LLMs with information retrieval techniques. 
Although RAG systems have proven effective, they still exhibit hallucinations \citep{magesh2024lhallucinationfree,niu-etal-2024-ragtruth-acl}. This underscores the need for rigorous evaluation before introducing such systems to users, especially within expert domains.

Evaluating LLMs on domain-specific benchmark datasets is thus essential for both research and industrial applications.
To support this, several evaluation frameworks--such as Language Model Evaluation Harness~\citep{gao2024eval-harness}, Holistic Evaluation of Language Models~\citep{liang2023holistic}--have been developed and widely adopted by the research community. 

Despite these developments, there remains a significant gap in the availability of comprehensive evaluation tools tailored for RAG pipelines, where multiple components influence overall accuracy: (1) retrieval corpus, (2) retrieval algorithms, (3) rerankers, (4) LLM backbones, and (5) evaluation metrics. For instance, \citet{magesh2024lhallucinationfree} shows 40-50\% of hallucinations can originate from the failure in document retrieval steps.

While there are existing tools \citep{rau-etal-2024-bergen-femnlp,zhang-etal-2024-raglab-emnlp-demo}, their utility is often limited to general benchmark datasets and corpora, such as MMLU \citep{hendryckstest2021mmlu} and Wikipedia while extending to other domain is not straightforward. Additionally, domain experts may struggle to adapt these tools to their specific goals, as many do not provide a 
 graphical interface (GUI).

Here we propose \ours\footnote{\Ours}, a holistic evaluation tool explicitly designed for assessing RAG systems in the legal domain. \ours\ extends Language Model Evaluation Harness \citep{gao2024eval-harness} by integrating it with \texttt{pyserini} \citep{Lin_etal_SIGIR2021_Pyserini} for information retrieval, while allowing easy control over individual components.
Furthermore, \ours\ supports the use of legal-specific corpora, such as Pile-of-Law \citep{henderson2022pileoflaw}, and benchmarks like LegalBench \citep{guha2023legalbench}, in an off-the-shelf manner. 
By providing a user-friendly GUI, \ours\ not only streamlines the evaluation process for legal researchers working on RAG but also enables legal AI practitioners to efficiently assess their models using domain-specific data. 

In summary, our contributions are as follows.
\begin{itemize}
    \item We propose \ours, an open-source evaluation tool for RAG systems that allows seamless integration of new corpus, tasks, and retrieval components.
    \item \ours\ features a user-friendly GUI, making it accessibility to domain experts.
    \item \ours\ provides pre-configured legal datasets for conducting RAG experiments with ease.
\end{itemize}

\section{Related work}

\begin{figure*}[tb]
    \centering
    \includegraphics[width=0.95\textwidth]{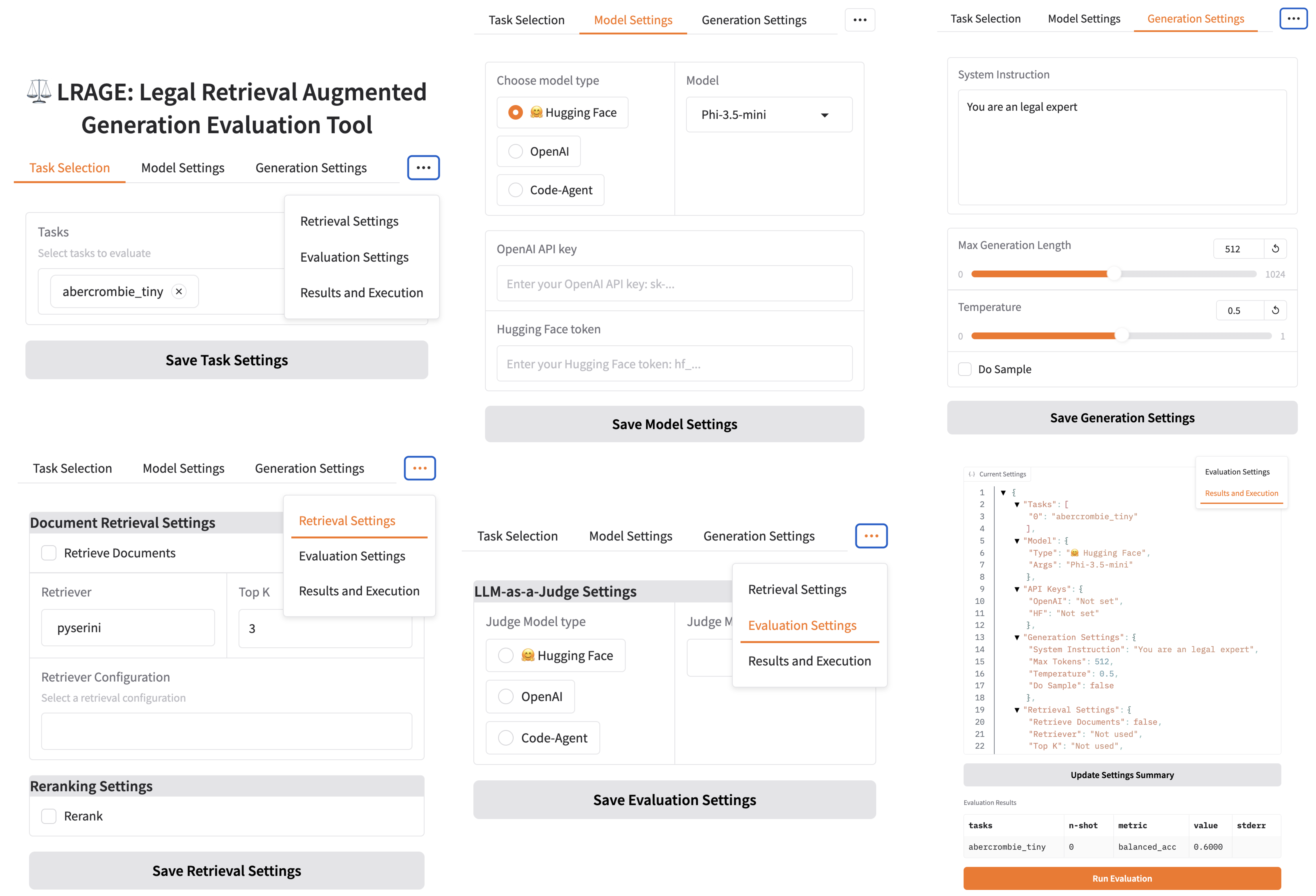}
    \caption{GUI of \ours. It consists of six tabs: Task (top-left), Model (top-center), Generation Parameters (top-right), Retriever (bottom-left), LLM-as-a-Judge (bottom-center), and a result tab(bottom-right). Each configuration tab allows users to define settings, which are then used in the final tab to perform experiments and immediately view the results.}
    \label{fig: gui}
\end{figure*}

\subsection{Legal Case Retrieval}
Finding relevant previous cases is critical for legal decision-making~\citep{feng2024acl-legal-case-retrieval-survey}. 
Accordingly, various studies have proposed models and datasets to address legal retrieval tasks~\citep{goebel2023coliee,santosh2024ecthrpcr,hou2024clercdatasetlegalcase,li2023sailer,li2023lecardv2largescalechineselegal,gao2024acl-enhancing-legal,zheng2025legalrag}.
However, no comprehensive evaluation tools have been developed to specialized in how retrieval performance in legal RAG systems is influenced by the choice of (1) retrieval corpus, (2) retreiver, (3) backbone LLMs, (4) reranker, and (5) rubric.

\subsection{RAG in legal domain}
\citet{magesh2024lhallucinationfree} analyzed commercial RAG systems in the U.S. legal domain using 202 examples, revealing that even the most competent system exhibited 17\% hallucination rate.
\citet{niu-etal-2024-ragtruth-acl} introduced RAGTruth benchmark, built using a subset of LegalBench \citep{guha2023legalbench}. Their evaluation is limited to the retrieval tasks.

\citet{zheng2025legalrag} developes two retrieval and RAG legal benchmarks: Bar Exam QA, and Housing Statute QA based on U.S. precedents and the statutory housing law. 
They show BM25 and current dense retrievers exhibit limited performance in recognizing gold passages in the legal domain.
Notably, they built large scale ground truth passages labeled by law students and legal experts.

\subsection{Legal Benchmarks for LLMs}
This section briefly reviews legal benchmarks designed for evaluating LLMs.
\citet{guha2023legalbench} proposed LegalBench, a benchmark comprising 162 legal language understanding tasks. These tasks are organized according to six types of legal reasoning based on the IRAC framework. LegalBench focuses exclusively on English legal language understanding.
\citet{kimyeeun2024femnlp-kbl} developed KBL, a benchmark dedicated to Korean legal language understanding. In addition to examples, they also provide resources for RAG experiments, including a corpus of Korean statutes and precedents \citet{hwang2022lboxopen}. 
\citet{fei-etal-2024-lawbench} developed LawBench comprising 20 Chinese legal tasks categorized into three levels--Memorization, Understanding, and Applying--based on Bloom’s taxonomy.

Except for KBL, these studies evaluated LLMs without incorporating RAG. Also, the KBL benchmark utilized only a basic RAG setup, employing a BM25 retriever without a reranker.

\subsection{RAG Evaluation Tools}

\citet{rau-etal-2024-bergen-femnlp} developed BERGEN, a tool designed for the systematic evaluation of RAG systems in question-answering (QA) tasks. BERGEN enables users to analyze the impact of individual system components,  offering comprehensive support for various retrievers, rerankers, and language models. It employs an abstract class architecture with YAML configurations, allowing users to extend and customize components according to their requirements. Additionally, BERGEN supports multilingual capability by providing Wikipedia indices in 12 languages and offering multilingual versions of benchmark datasets.

\citet{zhang-etal-2024-raglab-emnlp-demo} introduced RAGLAB, a RAG evaluation tool focused on the comparative analysis of different RAG algorithms rather than the individual pipeline components such as retrievers, rerankers, and LLMs. The framework provides six major RAG algorithms and incorporates  ten QA benchmarks, using Wikipedia as the retrieval corpus. RAGLAB also supports advanced evaluation metrics, such as ALCE 
 \citep{gao2023enabling-alce-emnlp} and FactScore \citep{min-etal-2023-factscore-emnlp}, for assessing generative tasks. Moreover, the frameworks allows researchers to easily integrate new RAG algorithms.

Despite their strengths, these RAG evaluation frameworks have some notable limitations. First, these frameworks primarily rely on Wikipedia as the sole retrieval source. In domains where specialized knowledge and domain-specific documentation are critical (e.g., legal or medical fields), this reliance is a significant limitation, as current frameworks fail to adequately address specialized retrieval scenarios.
Second, while BERGEN provides extensibility across various pipeline components, its flexibility comes at the cost of cumbersome setup process, often requiring complex code implementations and configuration files. This lack of a no-code evaluation  environment limits its accessibility and ease of use, particularly for domain experts.
In contrast, \ours\ is designed to address these gaps. It offers seamless integration with other retrieval corpora and tasks,  a user-friendly GUI for easy accessibility by domain experts, and off-the-shelf legal-specific datasets, as detailed in the next section.

\section{System}

\ours\ is built on top of the open-source Language Model Evaluation Tool, \texttt{lm\allowbreak -\allowbreak evaluation\allowbreak -\allowbreak harness} \citep{gao2024eval-harness}, by incorporating Retriever and Reranker modules for RAG.  
This allows us to inherit advantages such as extensibility to various task and models and flexible system instruction prompt tuning.

To support various retriever and reranker frameworks and models, \ours\ employs a modular architecture for these components (Fig. \ref{fig: system}). The system follows SOLID design principles, particularly leveraging dependency injection to ensure loosely coupling between components, thereby enabling high modularity and extensibility. This modular design facilitates straightforward integration of new models and datasets without modifying the core architecture. It defines abstract classes that specify the essential 
 operations for Retriever and Reranker in constructing the RAG pipeline, which can be implemented at either the framework or model level.

\subsection{Retreiver and Reranker Modules}
The Retriever module is currently implemented using 
 \texttt{pyserini} \citep{Lin_etal_SIGIR2021_Pyserini}, while the Reranker module utilizes \texttt{rerankers} \citep{clavie2024rerankers}. Both frameworks are highly flexible and support various  models. By modularizing these components at the framework level, \ours\ significantly reduces the implementation overhead typically required to support multiple models.

\subsection{Metric Module}
To support the evaluation of generative tasks, we extend the metric module of \texttt{lm\allowbreak -\allowbreak evaluation\allowbreak -\allowbreak harness} by integrating a custom LLM-as-a-Judge functionality. This enables flexible, rubric-based evaluation of legal benchmarks. By allowing rubrics to be defined at the instance level \citep{min-etal-2023-factscore-emnlp,kim2024biggenbenchprincipledbenchmark}, \ours\ facilitates more detailed and precise evaluations.
The users also can easily access the evaluation results through aggregated final scores while retaining the ability to review detailed instance-level rubric-based assessments via stored sample logs. The module leverages the existing LM class architecture of \texttt{lm\allowbreak -\allowbreak evaluation\allowbreak -\allowbreak harness}, ensuring compatibility with various frameworks and models that can serve as judges.

\subsection{Legal domain specialization}

\ours\ offers pre-configured settings for evaluating RAG systems in the legal domain.
It currently supports various legal benchmarks such as KBL~\citep{kimyeeun2024femnlp-kbl}, LegalBench~\citep{guha2023legalbench}, and LawBench~\citep{fei-etal-2024-lawbench}.
Beyond benchmark support, \ours\ provides preprocessed resources for legal copora such as Pile-of-Law, including a chunked version, a pre-compiled BM25 index, and a pre-compiled FAISS index \citep{douze2024faiss}, facilitating immediate use of legal datasets in RAG experiments.
The demo video for our system is available at \url{https://github.com/hoorangyee/LRAGE}.
\section{Experiments}

We used Llama-3.1-8B~\citep{grattafiori2024llama3herdmodels}, GPT-4~\citep{openai2023gpt4} and various other LLMs during our evaluations.
The Pile-of-Law~\citep{henderson2022pileoflaw} corpus was chunked 
in a similar manner to that described in~\citep{hou2024clercdatasetlegalcase}.
We converted the CAIL~\citep{xiao2018cail2018} training set into a retrieval corpus by concatenating the fact section with metadata.
For both CAIL and Korean precedents and statutes corpora~\citep{hwang2022lboxopen, kimyeeun2024femnlp-kbl}, we treated each individual judgment as a single document for indexing, following the setup in previous work~\citep{kimyeeun2024femnlp-kbl}.
Unless otherwise specified, we used Llama-3.1-8B, BM25, and the top 3 retrieved documents as the default setting for RAG experiments.

\section{Results}

To demonstrate \ours, we measured the overall performance of RAG systems on legal Benchmarks while varying the following components: retrieval corpus, retrieval algorithm, LLM backbones, and reranker.

\begin{table}[t]
\centering
\caption{Evaluation result on 2024 Korean Bar Exam subtasks from KBL 
 Benchmark~\citep{kimyeeun2024femnlp-kbl}.
Korean precedents and statues~\citep{hwang2022lboxopen, kimyeeun2024femnlp-kbl} (KoPS) or Korean wikipedia (kowiki) were used as the retrieval corpus.
The values in parentheses indicate the difference (in percentage points) compared to the score obtained without RAG.
}
\resizebox{\linewidth}{!}{
\begin{tabular}{l|ccc}
\toprule
Acc (\%, $\uparrow$) & civil & public & criminal \\
\midrule
\multicolumn{4}{c}{\textbf{Llama-3.1-8B-chat}} \\
\midrule
w/o RAG & 27.1 & 17.5 & \textbf{27.5} \\
\midrule
\multicolumn{4}{l}{\textbf{RAG w/ KoPS}} \\
\hline
BM25 & 28.6 (+1.5) & 35.0 (+17.5) & 12.5 (-15.0) \\
+ ColBERT reranker & 27.1 (+0.0) & 30.0 (+12.5) & 17.5 (-10.0) \\
+ T5 reranker & 31.4 (+4.3) & 35.0 (+17.5) & 15.0 (-12.5) \\
+ Cross-Encoder reranker & \textbf{31.4 (+4.3)} & \textbf{40.0 (+22.5)} & 17.5 (-10.0) \\
\hline
mE5-L Dense Retriever$^a$ & 21.4 (-5.7) & 27.5 (+10.0) & 15.0 (-12.5) \\
bge-m3 Dense Retriever$^b$ & 15.7 (-11.4) &	27.5 (+10.0) & 15.0 (-12.5) \\
\midrule
\multicolumn{4}{l}{\textbf{RAG w/ kowiki}} \\
\hline
BM25 & 27.1 (+0.0) & 27.5 (+10.0) & 15.0 (-12.5) \\
+ ColBERT reranker & 27.1 (+0.0) & 27.5 (+10.0) & 17.5 (-10.0) \\
+ T5 reranker & 31.4 (+4.3) & 25.0 (+7.5) & 15.0 (-12.5) \\
+ Cross-Encoder reranker & 27.1 (+0.0) & 17.5 (+0.0) & 17.5 (-10.0) \\
\bottomrule
\multicolumn{4}{c}{\textbf{GPT-4o}} \\
\midrule
w/o RAG & 44.3 & \textbf{57.5} & 32.5 \\
BM25 w/ KoPS & \textbf{57.1 (+12.8)} & 55.0 (-2.5) & \textbf{50.0 (+17.5)} \\
\bottomrule

\end{tabular}
}
\label{tbl_kbl}
\begin{tablenotes}[]
\footnotesize

\item $a$: \citet{wang2024multilinguale5textembeddings} ~~ $b$: \citet{chen2024bgem3embeddingmultilingualmultifunctionality}
\end{tablenotes}
\end{table}
\begin{table}[t]
\centering
\caption{LegalBench~\citep{guha2023legalbench} evaluation result. We adopted wiki and the subsets of Pile of Law (PoL)~\citep{henderson2022pileoflaw} for the retrieval corpus. 
PoL-cases includes "courtlistener\_opinions", "tax\_rulings", "canadian\_decisions", and "echr".
PoL-study incudes "cc\_casebooks".
The values in parentheses indicate the difference (in percentage points) compared to w/o RAG.}
\resizebox{\linewidth}{!}{
\begin{tabular}{l|ccc}
\toprule
\multirow{3}{*}{Acc (\%, $\uparrow$)}  & international & nys & personal \\
 & citizenship & judicial & jurisdiction \\
 & questions & ethics & \\
\midrule
w/o RAG & 51.3 & 69.4 & 54.7 \\
\midrule
wiki & \textbf{59.4 (+8.1)} & 68.7 (-0.7) & 59.9 (+5.2) \\
PoL-cases & 55.5 (+4.2) & \textbf{70.9 (+1.5)} & 56.2 (-1.5) \\
PoL-study-materials & 51.3 (+0.0) & 69.9 (+0.5) & \textbf{66.1 (+11.4)} \\
\bottomrule
\end{tabular}
}
\label{tbl_legalbench}
\end{table}
\begin{table}[t]
\scriptsize
\centering
\caption{LawBench~\citep{fei-etal-2024-lawbench} evaluation result. 
Three knowledge-intensive subtasks were evaluated here.
1-2: Knowledge Question Answering; 3-3: Charge Prediction; 3-4: Preson Term Prediction w.o. Article. We adopted Chinese Wikipedia (zhwiki) and the CAIL~\citep{xiao2018cail2018} train set for the retrieval corpus.
The values in parentheses indicate the difference (in percentage points) compared to the score obtained without RAG.
}
\begin{tabular}{l|ccc}
\toprule
\multirow{2}{*}{LawBench}  & 1-2 & 3-3 & 3-4 \\
 & ACC (\%, $\uparrow$) & F1 (\%, $\uparrow$) & -log distance ($\uparrow$) \\

\midrule
w/o RAG & 34.4 & 27.2 & 0.59 \\
\midrule
CAIL & \textbf{34.8 (+0.4)} & \textbf{41.2 (+14.0)} & \textbf{0.72 (+0.13)} \\
zhwiki & 31.8 (-2.6) & 22.8 (-4.4) & 0.54 (-0.05) \\
\bottomrule
\end{tabular}
\label{tbl_lawbench}
\end{table}

\paragraph{Retrieval corpus}

We first evaluate RAG performance on the multiple-choice questions from Korean Bar Exam using Llama-3.1-8B and GPT-4o\footnote{gpt-4o-2024-11-20} with \ours.
With Korean precedents and statues (KoPS) as the retrieval corpus, Llama shows improved performance compared to no RAG setting in \texttt{civil} and \texttt{public} subtasks (Table \ref{tbl_kbl}, 3rd vs. 5th--8th rows).
In contrast, using kowiki corpus yields little to no improvement, or in some cases results in lower performance (3rd vs. 12th--15th rows).
We also conducted RAG evaluation for LegalBench~\citep{guha2023legalbench} (English), and LawBench~\citep{fei-etal-2024-lawbench} (Chinese).
Table~\ref{tbl_legalbench} demonstrates that, on three selected knowledge-intensive subtasks from LegalBench, RAG generally improves accuracy (see diagonal entries in 3rd–5th rows), although the effectiveness still depends on the choice of retrieval corpus.
Similarly, evaluation on three subtasks from LawBench reveals corpus-dependent performance (Table~\ref{tbl_lawbench}).
These results align with the intuition that selecting a task-specific corpus is crucial for effective RAG.

\paragraph{Retrieval algorithms}
In the Korean Bar Exam experiments, the dense retriever underperforms compared to BM25 (Table~\ref{tbl_kbl}, 9th and 10th rows). 
This suggests that domain adaptation of dense retreivers is critical in the legal domain consistent with findings from recent studies \citet{zheng2025legalrag, hou2024clercdatasetlegalcase}.

\paragraph{Reranker}
Next, we examine how performance varies with the choice of rerankers.
Evaluation on the \textit{civil} and \textit{public} subtasks shows that  the cross-encoder reranker achieves the best performance in both cases (Table~\ref{tbl_kbl}, 8th row). 
However, this trend does not hold for the \textit{kowiki} corpus, where the T5-based reranker performs better on the \textit{civil} subtask. This demonstrates that reranker 
 effectiveness varies not only by task but also by corpus, highlighting the importance of the evaluation tools like \ours\ which allows seamless exploration of different RAG components combinations.

\paragraph{LLM backbones}
Interestingly, for \texttt{criminal} category, both KoPS and kowiki corpora show a significant drop in accuracy \textit{criminal} task (Table \ref{tbl_kbl}, 3rd vs 5th--8th and 12th--15th columns). This contrasts with the improvements observed in stronger API model (final two rows), suggesting that the base capability of the model has a substantial impact on RAG performance, again emphasizing the importance of evaluation tools like \ours.

Further experiments with other models and subtasks are presented in Appendix, demonstrating that the performance of a RAG system in the legal domain depends on multiple interacting  components.

\begin{table}[t]
\centering
\scriptsize 
\caption{LLM-as-a-judge score on PLAT with different rubric settings. 
$S_{\text{sem,struc}}$ and $S_\text{struc}$ stand for "semantic and structural rubrics" and "structural rubrics". The average scores from three independent experiments are shown.
GPT-4o was used as the judge model.}
\label{tbl_plat}
\setlength{\tabcolsep}{5pt}
\begin{tabular}{l|c c}
\hline
\textbf{Model} & \textbf{$S_{\text{sem,struc}}$} & \textbf{$S_\text{struc}$} \\
\hline
GPT-4o-mini & 4.16(±0.02) & 4.41(±0.07)  \\
GPT-4o & 4.26(±0.06) & 4.47(±0.07) \\
\hline
\end{tabular}
\end{table}
\paragraph{Rubrics}
We demonstrate the LLM-as-a-judge functionality using PLAT~\citep{choi2025taxationperspectiveslargelanguage}, a Korean taxation benchmark.

We first convert 50 yes/no questions about the legitimacy of additional tax penalties from PLAT into descriptive questions. To prepare the rubrics, we use the reasoning section of Korean precedents and automatically convert them into rubrics using GPT-o1. After manually revising 10 examples in  collaborating with a tax expert, we label the remaining 40 examples using GPT-o1 with a few-shot learning approach. More details will be provided in the paper currently in preparation.

The resulting PLAT rubrics consist of two types of items: (1) semantic and (2) structural. Semantic items evaluate the correctness of specific legal reasoning (e.g. "Is \texttt{[question-specific-article]} appropriately cited?"), while structural items focus on general aspects of writing (e.g. "Is the answer written concisely without unnecessary repetition?"). Each question includes four or five items, with a total possible score of 5 points. See Appendix for more examples.

To investigate how the scores depend on the choice of rubrics, we prepare a new set consisting only of structural items. Since structural understanding may not require deep legal knowledge or reasoning skills, less capable LLMs may achieve similar scores with more competent LLMs.
The results show that on the original rubrics, GPT-4o-mini achieves -0.1 score compared to GPT-4o (Table~\ref{tbl_plat}, 1st column), whereas the gap narrows to -0.06 on the new structural rubrics (final column).

We conduct additional experiments using the BigLaw Bench core samples, an English legal task dataset.\footnote{We used examples from \url{https://github.com/harveyai/biglaw-bench/blob/main/blb-core/core-samples.csv}.} The rubrics comprise sixty-four 1-point items and five 2-point items. We evaluate answers generated by GPT-4o-mini or GPT-4o using GPT-4o as a judge, which yields a mean score 5.63$\pm$0.45 (from three independent experiments). 
When the point values are swapped--1-point items changed to 2-points and vice versa--the mean score adjusts to 5.80$\pm$0.33. 
Although this indicate a rise in the average scores, it is difficult to draw a definitive  conclusions due to the limited number of examples (five). Nevertheless, combined with the earlier experiment using PLAT, the result highlights the importance of supporting rubric modifications when evaluating free-form text.

\begin{table}[t]
\centering
\scriptsize
\caption{Evaluation results of Bar Exam QA~\citep{zheng2025legalrag} with agentic RAG~\citep{smolagents2025}.}
\label{tbl_stanford_barexam_qa} 
\begin{tabular}{l|c c c}
\hline
\textbf{Model} & \textbf{w/o RAG} & \textbf{w/ BM25} & \textbf{w/ Agentic RAG} \\
\hline
Llama-3.1-8B-Instruct & 47.0 & 41.0 (-6.0) & 47.9 (+0.9) \\
Llama-3.1-70B-Instruct & 76.1 & 68.4 (-7.7) & 61.5 (-14.6)  \\
GPT-4o-mini & 48.7 & 51.3 (+2.6) & 51.3 (+2.6) \\  
\hline
\end{tabular}
\end{table}

\paragraph{Agentic RAG}
To support agentic RAG, \ours\ integrates \texttt{smolagents}~\citep{smolagents2025}. 
For the demonstration, we use recent legal RAG benchmark from \citet{zheng2025legalrag}. 
The results show that the off-the-shelf application of agentic RAG does not necessarily improve performance (Table~\ref{tbl_stanford_barexam_qa}, 2nd vs 3rd columns), although stronger model shows relatively more competent results (1st vs 2nd rows). This suggests that, similar to retrievers, domain adaptation of agent components--such as prompts, tools, and reasoning frameworks \cite{kang2023-chatgpt-irac,anthropic2025think_tool}--may be necessary.

\section{Conclusion}
We propose \ours, a holistic evaluation tool for RAG systems specifically tailored for applications in the legal domain.
Building on the widely adapted open-source LLM evaluation tools \texttt{lm-evaluation-harness}, 
\ours\ integrates two core functionalities--Retriever, Reranker--along with additional features for evaluating generative tasks using instance-level custom rubrics. Experiments on legal domain benchmakrs demonstrate how the overall performance of RAG systems depends on individual components, highlighting the effectiveness of \ours, which enables analysis with just a few lines of script or GUI. The inclusion of a user-friendly GUI and pre-processed legal corpora for retrieval facilitates seamless adaptation by legal domain experts, making \ours\ highly accessible and practical for specialized use cases.

\bibliography{legal_ai}

\FloatBarrier
\onecolumn
\newpage

\twocolumn
\appendix

\section{Additional Experiments} \label{sec:app_ee}

\subsection{KBL}

Here, we provides additional experimental results. Table~\ref{kbl_knowledge_rag} shows the results of the KBL Legal Knowledge benchmark conducted using the Llama-3.1-8B model. Similar to the result in Table \ref{tbl_kbl}, retrieving 5 documents from KoPS consistently yields better performance compared to the case of \texttt{kowiki}.

\begin{table}[htbp]
\scriptsize  
\centering
\caption{Additional evaluation result on Legal Knowledge subtasks from KBL Benchmark~\citep{kimyeeun2024femnlp-kbl}. Llama-3.1-8B and BM25 retriever were used. The number indicates the average scores over 7 subtasks.  }
\begin{tabular}{l|l|cc}
\toprule
\textbf{Corpus} & \textbf{Reranker} & \textbf{Top-k=3} & \textbf{Top-k=5} \\
\midrule
\multicolumn{2}{c|}{w/o RAG} & \multicolumn{2}{c}{23.9} \\
\midrule
\multirow{4}{*}{KoPS} & - & 31.0 & 33.5 \\
 & colbert        & 31.6 & 33.0 \\
 & cross\_encoder & 29.4 & 31.8 \\
 & t5             & 30.0 & 31.6 \\
\midrule
\multirow{4}{*}{kowiki} & - & 15.1 & 14.5 \\
 & colbert        & 15.1 & 14.5 \\
 & cross\_encoder & 14.9 & 14.9 \\
 & t5             & 14.3 & 15.7 \\
\hline
\end{tabular}
\label{kbl_knowledge_rag}
\end{table}

Table~\ref{tbl_kbl_appendix} presents the 2025 Korean Bar Exam results with additional models. 
\begin{table}[h!]
\centering
\scriptsize
\caption{Evaluation results w/o RAG on 2024 Korean Bar Exam from KBL Benchmark~\citep{kimyeeun2024femnlp-kbl} with more various models.}
\label{tbl_kbl_appendix}
\begin{tabular}{l|ccc}
\toprule
Acc (\%, $\uparrow$) & civil & public & criminal \\
\midrule
Gemma-3-4b-it$^{a}$ & 37.1 & 22.5 & 25.0 \\
Gemma-3-12b-it$^{a}$ & 28.6 & 35.0 & 42.5 \\
EXAONE-3.0-7.8B-Instruct$^{b, *}$ & 20.0 & 20.0 & 22.5 \\
GPT-4o-mini-2024-07-18$^{*}$ &  31.4 & 32.5 & 25.0 \\
\bottomrule
\end{tabular}

\smallskip
\scriptsize\raggedright
$^{a}$\citet{gemmateam2025gemma3technicalreport}.~~~$^{b}$\citet{research2024exaone3078binstruction}.
\\$^{*}$Results cited from \citet{kimyeeun2024femnlp-kbl}.
\end{table}

\subsection{LegalBench}
Here we present our initial experiments on LegalBench. Table~\ref{tbl_acc_model_corpus} presents the results of RAG experiments on LegalBench-tiny with Pile-of-Law-mini corpus.
LegalBench Tiny is a subset of LegalBench~\citep{guha2023legalbench}, constructed by randomly sampling 10 instances per subtask, with sampling stratified to ensure a balanced distribution of correct answers. 
Pile-of-Law-mini corpus consists of 10\% of randomly sampled documents from the original corpus. 

\paragraph{Retrieval corpus}
We first evaluate RAG systems while varying their retrieval corpus: (1) no retrieval (Table \ref{tbl_acc_model_corpus} 1st panel), (2) Wikipedia (2nd panel), and (3) Pile-of-Law-mini (3rd panel). The result highlight the clear importance of using domain specific legal corpus. Interestingly, the accuracy of GPT-4o-mini decreases the most with Pile-ofLaw-mini (3rd panel, 4th row). To investigate this, we altered the input order from instruction + retrieved-documents + examples + questions to retrieved-documents + instruction + examples + questions and observed an increase in accuracy (indicated by diff. prompt, final row of each panel). We suspect this behavior is due to the unique structure of legal documents and GPT-4o-mini’s limited adaptation to the legal domain.

\paragraph{LLM backbones}
Next we demonstrate how the choice of LLM backbones, which generate the final answers, impacts performance (Table \ref{tbl_acc_model_corpus}). The results reveals significant variance between models.

\begingroup
\setlength{\tabcolsep}{0.5pt}
\renewcommand{\arraystretch}{1}
\begin{table}[ht]
\centering
\scriptsize 
\caption{Evaluation results of LegalBench-Tiny.}
\label{tbl_acc_model_corpus}
\begin{tabular}{lccccc}
\toprule
\textbf{Model} & \textbf{Avg} & \textbf{Interpretation} & \textbf{Issue} & \textbf{Rhetorical} & \textbf{Rule} \\
\midrule

\multicolumn{6}{c}{w/o RAG} \\
\hline
Llama3.1-8B & 66.7 & 68.7 & 64.6 & 65.6 & 67.9 \\
Qwen2.5-7B & 67.7 & 72.5 & 70.5 & 62.9 & 64.7 \\
SaulLM-7B & 57.4 & 60.7 & 52.9 & 50.3 & 60.9 \\
\hline
GPT-4o-mini & 64.9 & 65.9 & 54.6 & 67.6 & 71.3 \\
+ diff. prompt & 72.7 & 75.2 & 73.3 & 74.2 & 67.9 \\
\midrule

\multicolumn{6}{c}{BM25 for Wikipedia} \\
\hline
Llama3.1-8B & 67.4 (+0.7) & 70.4 & 70.7 & 60.0 & 68.4 \\
Qwen2.5-7B & 66.3 (-1.4) & 72.1 & 70.1 & 60.5 & 62.3 \\
SaulLM-7B & 56.3 (-1.1) & 62.9 & 50.0 & 53.5 & 58.8 \\
\hline
GPT-4o-mini & 58.9 (-6.0) & 53.0 & 58.0 & 63.2 & 61.4 \\
+ diff. prompt & 71.4 (-1.3) & 75.3 & 72.1 & 66.9 & 71.1 \\
\midrule

\multicolumn{6}{c}{BM25 for Pile-of-Law-mini} \\
\hline
Llama3.1-8B & 68.1 (+1.4) & 69.6 & 71.5 & 63.5 & 67.8 \\
Qwen2.5-7B & 66.5 (-1.2) & 72.7 & 66.6 & 60.4 & 66.1 \\
SaulLM-7B & 58.2 (+0.8) & 63.3 & 52.9 & 55.6 & 61.1 \\
\hline
GPT-4o-mini & 55.6 (-9.3) & 50.8 & 55.8 & 50.1 & 65.5 \\
+ diff. prompt & 73.1 (+0.4) & 73.3 & 69.7 & 75.2 & 74.2 \\
\bottomrule
\end{tabular}
\end{table}
\endgroup

\paragraph{Retrieval algorithms}
Next, we examine the effect of the retrieval algorithm by replacing BM25 baseline with a dense retriever. We use LegalBERT-base \citep{chalkidis2020legalbert}, and LexLM-base \citep{chalkidis2023lexlms}, as  encoder backbone (Table \ref{tbl_acc_retriever_reranker}). 
Using original DPR, fine-tuend for general-domain tasks, we observed similar performance to BM25 (1st vs 2nd rows). However, with domain-specialized encoders, there was a significant improvement in accuracy (3rd and 4th rows). When dense retriever finetuned on legal retrieval tasks was used, performance increased further (5th row), consistent with previous findings \citep{hou2024clercdatasetlegalcase}.

We also evaluated the effect of introducing ColBERT-based reranker\citep{khattab2020colbert-sigir}. Interestingly, the ColBERT reranker did not improve performance. This result suggests that using a reranker trained in the general domain can reduce the accuracy of RAG system, algining with recent findings from \citep{pipitone2024legalbenchrag}.
\begingroup
\begingroup
\setlength{\tabcolsep}{0.5pt} 
\renewcommand{\arraystretch}{1} 

\begin{table}[h!]
\scriptsize
\centering
\caption{Performance table under varying retrieval algorithms (model fixed to GPT-4o-mini), The subset of Pile-of-Law is used as a retrieval pool.
LegalBERT-C, LegalBERT-CR, and LegalBERT-CR$_{GPT}$ Stand for "LegalBERT-DPR-CLERC", "LegalBERT-DPR-CLERC + Reranker", "LegalBERT-DPR-CLERC  + Reranker (GPT-4o)" respectively.
}
\label{tbl_acc_retriever_reranker}
\begin{threeparttable}

\begin{tabular}{l|c|cccc}

\hline
\textbf{Retrieval Algorithms} &\textbf{Avg} & \textbf{Interpretation} & \textbf{Issue} & \textbf{Rhetorical} & \textbf{Rule} \\
\hline
BM25  &55.6& 50.8 & 55.8 & 50.1 & 65.5 \\
DPR  & 55.1  & 54.7 & 53.4 & 45.6 & 66.7 \\
\hline 
LegalBERT  & 60.4  & 54.7 & 55.6 & 59.0 & 72.3 \\
LexLM-base  & 60.3  & 57.0 & 55.8 & 57.9 & 70.5 \\
LegalBERT-C  & 63.7  & 60.0 & 54.2 & 71.4 & 69.0 \\
\hline
BM25  + Reranker & 55.1  & 50.4 & 54.2 & 51.2 & 64.5 \\
LegalBERT-CR & 63.5  & 58.7 & 54.8 & 70.9 & 69.7 \\
LegalBERT-CR$_{GPT}$ & 63.5  & 58.4 & 58.2 & 67.5 & 69.7 \\
LexLM-base  + Reranker & 60.4  & 58.8 & 53.3 & 63.9 & 65.7 \\
\hline
\end{tabular}
  \end{threeparttable}
\end{table}

\endgroup

\paragraph{Experiments shown in Table \ref{tbl_legalbench}}
LegalBench also include non-knowledge-intensive subtasks where external documents are not required to answer the questions. Additionally, Pile-of-Law comprises a wide range of legal documents, many of which may not be directely relevant.
To better evaluate \ours\ in a more controlled setting and to enhance interpretability, we focus on the three knowledge-intensive subtasks from LegalBench and use all corresponding examples. Similarly, instead of random sampling, we construct subsets of Pile-of-Law by categorizing documents based on 
 type.

\subsection{LawBench}
Table~\ref{tbl_lawbench_appendix} presents additional results on LawBench for models not included in the main text. For InternLM2~\citep{cai2024internlm2technicalreport}, RAG improves performance in sections 3-3 and 3-4, but leads to lower scores in section 1-2. In contrast, for Qwen2.5-7B~\cite{qwen2.5}, RAG improves scores in section 1-2 but results in lower scores in sections 3-3 and 3-4.
\begin{table}[tb]
\scriptsize
\centering
\caption{Evaluation of additional LLMs on LawBench~\citep{fei-etal-2024-lawbench}.
Three knowledge-intensive subtasks were evaluated here.
1-2: Knowledge Question Answering; 3-3: Charge Prediction; 3-4: Preson Term Prediction w.o. Article. We adopted Chinese Wikipedia (zhwiki) and the CAIL~\citep{xiao2018cail2018} train set for the retrieval corpus.
}
\begin{tabular}{l|ccc}
\toprule
\multirow{2}{*}{LawBench}  & 1-2 & 3-3 & 3-4 \\
 & ACC (\%, $\uparrow$) & F1 (\%, $\uparrow$) & -log distance ($\uparrow$) \\
\midrule
\multicolumn{4}{c}{internlm2-chat-7b} \\
\midrule
w/o RAG & \textbf{39.4} & 50.0 & 62.1 \\
\midrule
CAIL & 24.6 & \textbf{52.0} & \textbf{74.5} \\
zhwiki & 25.8 & 48.0 & 69.0 \\
\midrule
\multicolumn{4}{c}{Qwen2.5-7B-Instruct-1M} \\
\midrule
w/o RAG & 29.4 & \textbf{56.0} & \textbf{75.0} \\
\midrule
CAIL & \textbf{54.5} & 48.4 & 64.7 \\
zhwiki & 45.0 & 43.8 & 65.7 \\
\bottomrule
\end{tabular}
\label{tbl_lawbench_appendix}
\end{table}

\subsection{PLAT}

\begin{table}[h]
\centering
\caption{Examples of rubrics used in the taxation dataset (PLAT)}
\label{tbl_plat_rubric}
\tiny
\begin{tabular}{>{\raggedright\arraybackslash}p{1.5cm} | >{\raggedright\arraybackslash}p{6cm}}
\toprule
\textbf{Rubric Type} & \multicolumn{1}{c}{\textbf{Content}} \\
\midrule
\midrule
Structural & \texttt{\detokenize{
"Below are 5 evaluation criteria (total 5 points) for the answer on 'The Legitimacy of the Penalty Tax Imposition' based on the above case and explanation. 1. Structure and length of the writing (1 point): Evaluates whether the writing follows a logical order (introduction-main-conclusion, etc.) and is written concisely without unnecessarily excessive length (verbose description). 2. Formal completeness (1 point): Evaluates whether paragraphs are divided according to the logical flow without unnecessarily verbose expressions. 3. Clarity of introduction and problem statement (1 point): Whether the facts given in the case are concisely summarized and the issue (legitimacy of penalty tax imposition) is clearly presented. 4. Accuracy of citing relevant laws and precedents (1 point): Evaluates whether the laws and precedents necessary for problem-solving such as Value Added Tax Act, Enforcement Decree, Enforcement Rules, Framework Act on National Taxes, precedents, etc. are appropriately cited and properly connected to the necessary parts. 5. Adherence to expression (1 point): Evaluates whether the case overview and the requirements of the problem are faithfully reflected."
}} \\
\midrule
Semantic and Structural & \texttt{\detokenize{
"Below are 5 evaluation criteria (total 5 points) for the answer on 'The Legitimacy of the Penalty Tax Imposition' based on the above case and explanation. Lower points are allocated to items evaluating form, while higher points are allocated to items evaluating content. 1. (Form) Structure and length of the writing (0.5 points) - Whether the writing follows a logical order (introduction-main-conclusion, etc.) - Whether it is written concisely without unnecessarily excessive length (verbose description) 2. (Content) Summary of facts and presentation of main issues (1 point) - Whether the facts appearing in the case are accurately identified and key issues are concisely presented - Whether it clearly emphasizes that the legitimacy of the penalty tax imposition is at issue 3. (Content) Appropriateness of relevant laws and interpretation (1 point) - Whether appropriate citations are made to the Corporate Tax Act (provisions regarding investment trusts being considered domestic corporations), Framework Act on National Taxes, or necessary tax law provisions - Whether it specifically explains how these provisions can/cannot be applied to impose penalty tax 4. (Content) Judgment of legitimate reasons and thoroughness of argumentation (1.5 points) - Whether the plaintiff's argument ('investment trust is not a taxpayer', 'excessive refund was inevitable') and the defendant's argument ('penalty tax imposition is justified for excessive refund application') are compared and examined - Whether the existence of 'legitimate reasons' that could excuse the plaintiff from negligence in the refund procedure is logically analyzed 5. (Content) Validity and clarity of conclusion presentation (1 point) - Whether a clear conclusion is drawn on whether the penalty tax imposition is legitimate or illegitimate - Whether the reasons supporting the conclusion (key issues and results of legal review) are presented concisely and clearly"}} \\
\bottomrule
\end{tabular}
\end{table}
Table~\ref{tbl_plat_rubric} presents the rubric used in the PLAT experiment. The content was machine-translated from Korean to English.

\end{document}